\documentclass{article}

\usepackage{times}
\usepackage{graphicx} 
\usepackage{subcaption}

\usepackage{natbib}

\usepackage{algorithm}
\usepackage{algorithmic}

\usepackage{hyperref}

\usepackage{amssymb} 
\usepackage{amsmath} 
\usepackage{float}
\usepackage{amsthm}
\DeclareMathOperator{\E}{\mathbb{E}}

\DeclareMathOperator{\tr}{tr}
\DeclareMathOperator{\diag}{diag}

\def\!#1{\boldsymbol{#1}}
\def\*#1{\mathbf{#1}}

\usepackage[accepted]{icmlnew}

\icmltitlerunning{Multiplicative Normalizing Flows for Variational Bayesian Neural Networks}

\begin{document} 

\twocolumn[
\icmltitle{Multiplicative Normalizing Flows for Variational Bayesian Neural Networks}



\icmlsetsymbol{equal}{*}

\begin{icmlauthorlist}
\icmlauthor{Christos Louizos}{to,tno}
\icmlauthor{Max Welling}{to,cifar}
\end{icmlauthorlist}

\icmlaffiliation{to}{University of Amsterdam, Netherlands}
\icmlaffiliation{tno}{TNO Intelligent Imaging, Netherlands}
\icmlaffiliation{cifar}{Canadian Institute For Advanced Research (CIFAR)}

\icmlcorrespondingauthor{Christos Louizos}{c.louizos@uva.nl}

\icmlkeywords{Bayesian neural networks, variational inference, deep learning, uncertainty quantification}

\vskip 0.3in
]



\printAffiliationsAndNotice{}  

\begin{abstract} 
We reinterpret multiplicative noise in neural networks as auxiliary random variables that augment the approximate posterior in a variational setting for Bayesian neural networks. We show that through this interpretation it is both efficient and straightforward to improve the approximation by employing normalizing flows~\cite{rezende2015variational} while still allowing for local reparametrizations~\citep{kingma2015variational} and a tractable lower bound~\cite{ranganath2015hierarchical,maaloe2016auxiliary}. In experiments we show that with this new approximation we can significantly improve upon classical mean field for Bayesian neural networks on both predictive accuracy as well as predictive uncertainty. 
\end{abstract} 

\section{Introduction}
\begin{figure*}[htb]
\centering
\begin{subfigure}{.5\textwidth}
  \centering
  \includegraphics[width=\linewidth,scale=0.1]{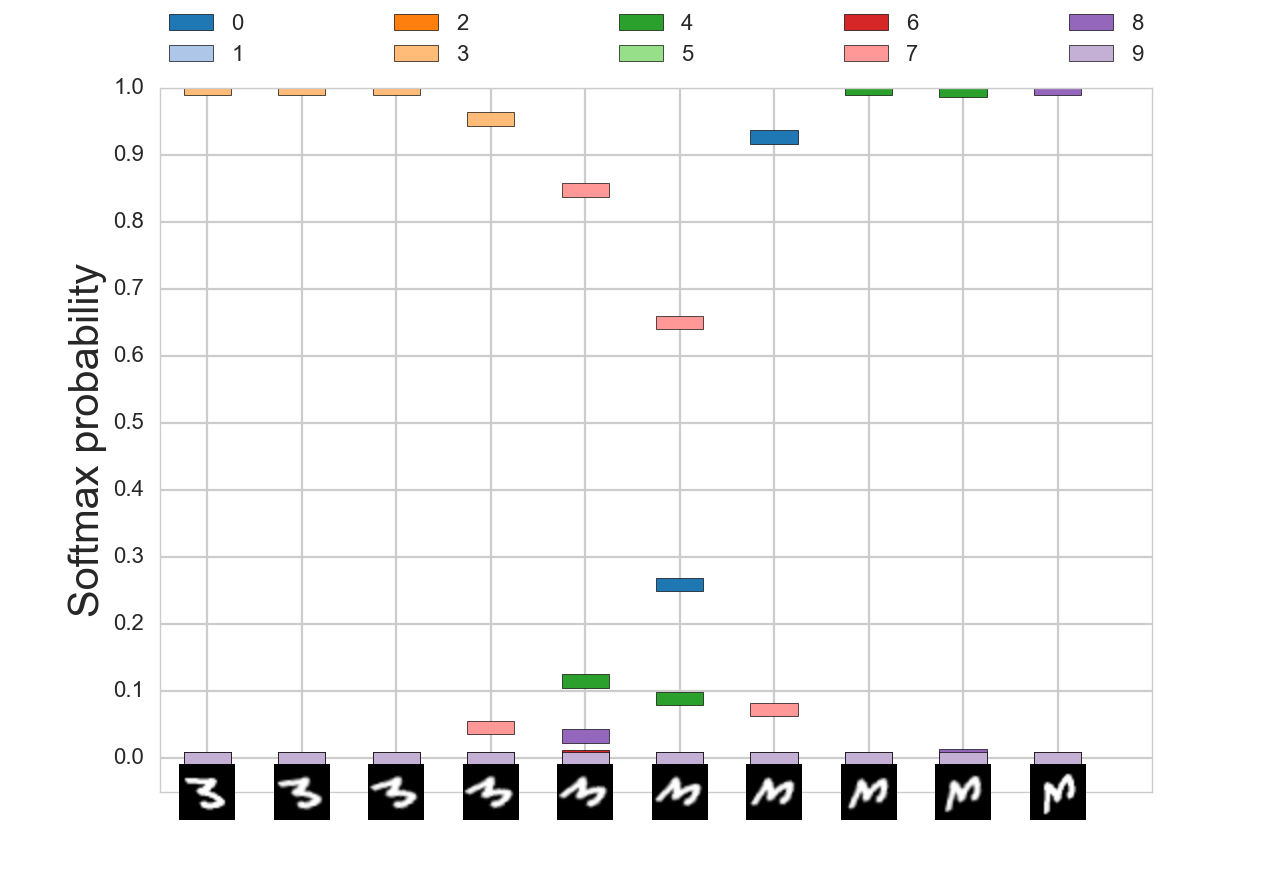}
  \caption{LeNet with weight decay}
  \label{fig:lenet_de_unc}
\end{subfigure}%
\begin{subfigure}{.5\textwidth}
  \centering
  \includegraphics[width=\linewidth, scale=0.1]{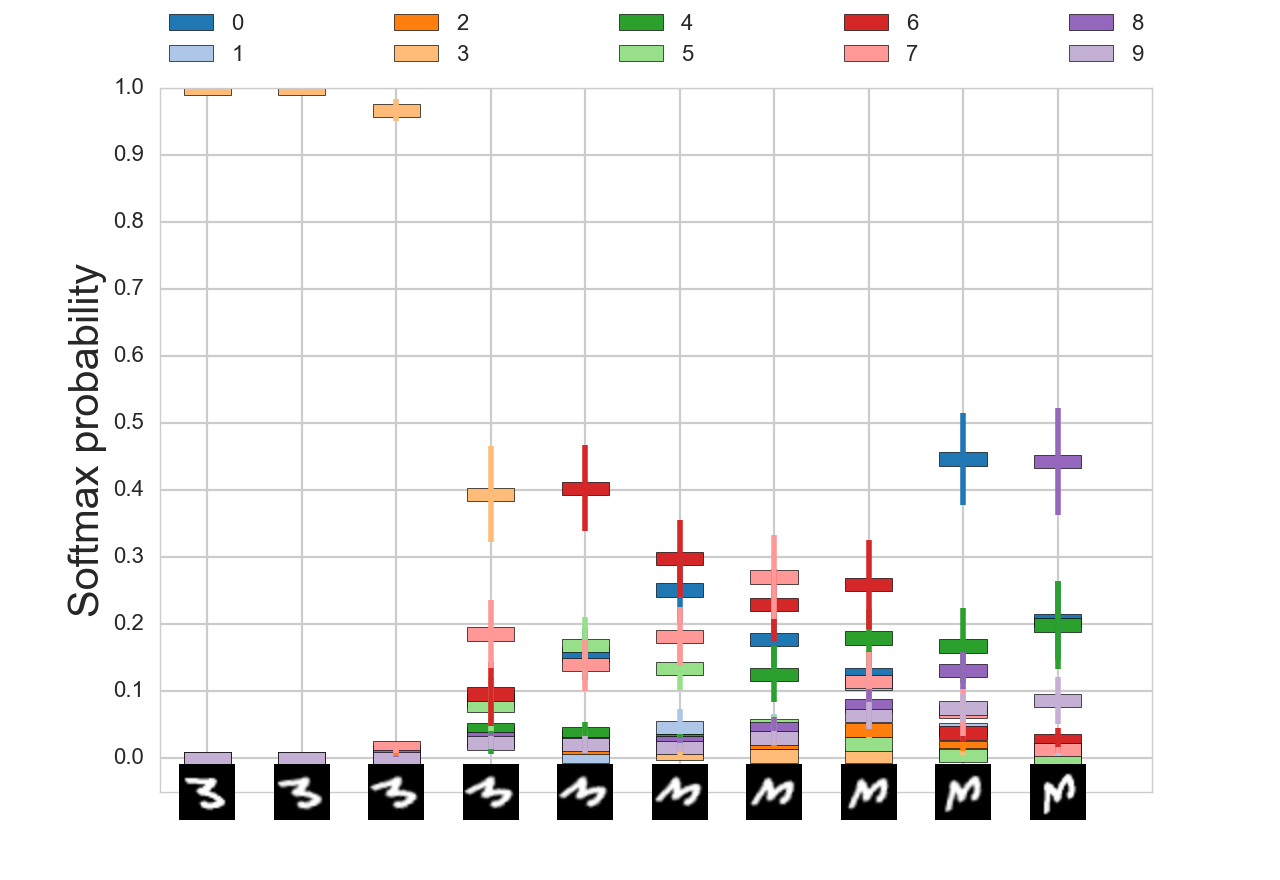}
  \caption{LeNet with multiplicative formalizing flows}
  \label{fig:lenet_mnf_unc}
\end{subfigure}
\caption{Predictive distribution for a continuously rotated version of a 3 from MNIST. Each colour corresponds to a different class and the height of the bar denotes the probability assigned to that particular class by the network. Visualization inspired by~\cite{gal2015dropout}.}
\label{fig:intro_fig}
\end{figure*}

Neural networks have been the driving force behind the success of deep learning applications. Given enough training data they are able to robustly model input-output relationships and as a result provide high predictive accuracy. However, they do have some drawbacks. In the absence of enough data they tend to overfit considerably; this restricts them from being applied in scenarios were labeled data are scarce, e.g. in medical applications such as MRI classification. Even more importantly, deep neural networks trained with maximum likelihood or MAP procedures tend to be overconfident and as a result do not provide accurate confidence intervals, particularly for inputs that are far from the training data distribution. A simple example can be seen at Figure~\ref{fig:lenet_de_unc}; the predictive distribution becomes overly overconfident, i.e. assigns a high softmax probability, towards the wrong class for things it hasn't seen before (e.g. an MNIST 3 rotated by 90 degrees). This in effect makes them unsuitable for applications where decisions are made, e.g. when a doctor determines the disease of a patient based on the output of such a network. 

A principled approach to address both of the aforementioned shortcomings is through a Bayesian inference procedure. Under this framework instead of doing a point estimate for the network parameters we infer a posterior distribution. These distributions capture the parameter uncertainty of the network, and by subsequently integrating over them we can obtain better uncertainties about the predictions of the model. We can see that this is indeed the case at Figure~\ref{fig:lenet_mnf_unc}; the confidence of the network for the unseen digits is drastically reduced when we are using a Bayesian model, thus resulting into more realistic predictive distributions. Obtaining the posterior distributions is however no easy task, as the nonlinear nature of neural networks makes the problem intractable. For this reason approximations have to be made.

Many works have considered the task of approximate Bayesian inference for neural networks using either Markov Chain Monte Carlo (MCMC) with Hamiltonian Dynamics~\citep{neal1995bayesian}, distilling SGD with Langevin Dynamics~\citep{welling2011bayesian,korattikara2015bayesian} or deterministic techniques such as the Laplace Approximation~\citep{mackay1992practical}, Expectation Propagation~\citep{hernandez2015probabilistic,hernandez2015black} and variational inference~\citep{graves2011practical,blundell2015weight,kingma2015variational,gal2015dropout,louizos2016structured}.

In this paper we will also tackle the problem of Bayesian inference in neural networks. We will adopt a stochastic gradient variational inference~\citep{kingma2013auto,rezende2014stochastic} procedure in order to estimate the posterior distribution over the weight matrices of the network. Arguably one of the most important ingredients of variational inference is the flexibility of the approximate posterior distribution; it determines how well we are able to capture the true posterior distribution and thus the true uncertainty of our models. In Section~\ref{model_derivation} we will show how we can produce very flexible distributions in an efficient way by employing auxiliary random variables~\cite{agakov2004auxiliary,salimans2013fixed,ranganath2015hierarchical,maaloe2016auxiliary} and normalizing flows~\cite{rezende2015variational}. In Section~\ref{related} we will discuss related work, whereas in Section~\ref{experiments} we will evaluate and discuss the proposed framework. Finally we will conclude with Section~\ref{conclusion}, where we will provide some final thoughts along with promising directions for future research.

\section{Multiplicative normalizing flows}
\label{model_derivation}
\subsection{Variational inference for Bayesian Neural Networks}
Let $\mathcal{D}$ be a dataset consisting of input output pairs $\{(\*x_1, \*y_1), \dots, (\*x_n, \*y_n)\}$ and let $\*W_{1:L}$ denote the weight matrices of $L$ layers. Assuming that $p(\*W_{i})$, $q_\phi(\*W_{i})$ are the prior and approximate posterior over the parameters of the $i$'th layer we can derive the following lower bound on the marginal log-likelihood of the dataset $\mathcal{D}$ using variational Bayes~\citep{peterson1987mean,hinton1993keeping,graves2011practical,blundell2015weight,kingma2015variational,gal2015dropout,louizos2016structured}:
\begin{align}
\mathcal{L(\phi)} & = \E_{q_\phi(\*W_{1:L})}\big[\log p(\*y | \*x, \*W_{1:L}) + \nonumber\\ 
 & + \log p(\*W_{1:L}) - \log q_\phi(\*W_{1:L})\big],
\end{align}
where $\tilde{p}(\*x, \*y)$ denotes the training data distribution and $\phi$ the parameters of the variational posterior. For continuous $q(\cdot)$ distributions that allow for the reparametrization trick~\citep{kingma2013auto} or stochastic backpropagation~\citep{rezende2014stochastic} we can reparametrize the random sampling from $q(\cdot)$ of the lower bound in terms of noise variables $\epsilon$ and deterministic functions $f(\phi, \epsilon)$:
\begin{align}
\mathcal{L} & = \E_{p(\epsilon)}\big[\log p(\*y | \*x, f(\phi, \epsilon)) + \nonumber \\ & + \log p(f(\phi, \epsilon)) - \log q_\phi(f(\phi, \epsilon))\big].
\end{align}
This reparametrization allow us to treat approximate parameter posterior inference as a straightforward optimization problem that can be optimized with off-the-shelf (stochastic) gradient ascent techniques.

\subsection{Improving the variational approximation}
For Bayesian neural networks the most common family for the approximate posterior is that of mean field with independent Gaussian distributions for each weight. Despite the fact that this leads to a straightforward lower bound for optimization, the approximation capability is quite limiting; it corresponds to just a unimodal ``bump" on the very high dimensional space of the parameters of the neural network. There have been attempts to improve upon this approximation with works such as~\cite{gal2015dropout} with mixtures of delta peaks and~\cite{louizos2016structured} with matrix Gaussians that allow for non-trivial covariances among the weights. Nevertheless, both of the aforementioned methods are still, in a sense, limited; the true parameter posterior is more complex than delta peaks or correlated Gaussians.

There has been a lot of recent work on ways to improve the posterior approximation in latent variable models with normalizing flows~\cite{rezende2015variational} and auxiliary random variables~\cite{agakov2004auxiliary,salimans2013fixed,ranganath2015hierarchical,maaloe2016auxiliary} being the most prominent. Briefly, a normalizing flow is constructed by introducing parametrized bijective transformations, with easy to compute Jacobians, to random variables with simple initial densities. By subsequently optimizing the parameters of the flow according to the lower bound they can significantly improve the posterior approximation. Auxiliary random variables instead construct more flexible distributions by introducing latent variables in the posterior itself, thus defining the approximate posterior as a mixture of simple distributions.

Nevertheless, applying these ideas to the parameters in a neural network has not yet been explored. While it is straightforward to apply normalizing flows to a sample of the weight matrix from $q(\*W)$, this quickly becomes very expensive; for example with planar flows~\cite{rezende2015variational} we will need two extra matrices for each step of the flow. Furthermore, by utilizing this procedure we also lose the benefits of local reparametrizations~\cite{kingma2015variational,louizos2016structured} which are possible with Gaussian approximate posteriors. 

In order to simultaneously maintain the benefits of local reparametrizations and increase the flexibility of the approximate posteriors in a Bayesian neural network we will rely on auxiliary random variables~\cite{agakov2004auxiliary,salimans2013fixed,salimans2015markov,ranganath2015hierarchical,maaloe2016auxiliary}; more specifically we will exploit the well known ``multiplicative noise" concept, e.g. as in (Gaussian) Dropout~\cite{srivastava2014dropout}, in neural networks and we will parametrize the approximate posterior with the following process:
\begin{align}
\*z \sim q_\phi(\*z); \qquad \*W \sim q_\phi(\*W|\*z), \label{eq:mixture}
\end{align}
where now the approximate posterior becomes a compound distribution, $q(\*W) = \int q(\*W|\*z) q(\*z) d\*z$, with $\*z$ being a vector of random variables distributed according to the mixing density $q(\*z)$. To allow for local reparametrizations we will parametrize the conditional distribution for the weights to be a fully factorized Gaussian. Therefore we assume the following form for the fully connected layers:
\begin{align}
q_\phi(\*W|\*z) = \prod_{i=1}^{D_{in}}\prod_{j=1}^{D_{out}}\mathcal{N}(z_i\mu_{ij}, \sigma^2_{ij}), \label{eq:base_density_ff}
\end{align}
where $D_{in}, D_{out}$ is the input and output dimensionality, and the following form for the kernels in convolutional networks: 
\begin{align}
q_\phi(\*W|\*z) = \prod_{i=1}^{D_{h}}\prod_{j=1}^{D_{w}}\prod_{k=1}^{D_{f}}\mathcal{N}(z_k\mu_{ijk}, \sigma^2_{ijk}),
\end{align}
where $D_{h}, D_{w}, D_{f}$ are the height, width and number of filters for each kernel. Note that we did not let $\*z$ affect the variance of the Gaussian approximation; in a pilot study we found that this parametrization was prone to local optima due to large variance gradients, an effect also observed with the multiplicative parametrization of the Gaussian posterior~\cite{kingma2015variational,2017arXiv170105369M}. We have now reduced the problem of increasing the flexibility of the approximate posterior over the weights $\*W$ to that of increasing the flexibility of the mixing density $q(\*z)$. Since $\*z$ is of much lower dimension, compared to $\*W$, it is now straightforward to apply normalizing flows to $q(\*z)$; in this way we can significantly enhance our approximation and allow for e.g. multimodality and nonlinear dependencies between the elements of the weight matrix. This will in turn better capture the properties of the true posterior distribution, thus leading to better performance and predictive uncertainties. We will coin the term \emph{multiplicative normalizing flows} (MNFs) for this family of approximate posteriors. Algorithms~\ref{alg:ff_bnn},~\ref{alg:conv_bnn} describe the forward pass using local reparametrizations for fully connected and convolutional layers with this type of approximate posterior.

\begin{algorithm}[htb]
\caption{Forward propagation for each fully connected layer $h$. $\*M_w, \!\Sigma_w$ are the means and variances of each layer, $\*H$ is a minibatch of activations  and $\text{NF}(\cdot)$ is the normalizing flow described at eq.~\ref{eq:nf_bnn}. For the first layer we have that $\*H = \*X$ where $\*X$ is the minibatch of inputs.} 
\label{alg:ff_bnn}
\begin{algorithmic}[1]
\REQUIRE $\*H, \*M_w, \!\Sigma_w$
    \STATE $\*Z_0 \sim q(\*z_0)$
    \STATE $\*Z_{T_f} = \text{NF}(\*Z_0)$
    \STATE $\*M_h = (\*H\odot\*Z_{T_f})\*M_w$
    \STATE $\*V_h = \*H^2\!\Sigma_w$
    \STATE $\*E \sim \mathcal{N}(0, 1)$
    \STATE return $\*M_h + \sqrt{\*V_h} \odot \*E$
\end{algorithmic}
\end{algorithm}

\begin{algorithm}[htb]
\caption{Forward propagation for each convolutional layer $h$. $N_f$ are the number of convolutional filters, $*$ is the convolution operator and we assume the [batch, height, width, feature maps] convention.} 
\label{alg:conv_bnn}
\begin{algorithmic}[1]
\REQUIRE $\*H, \*M_w, \!\Sigma_w$
    \STATE $\*z_0 \sim q(\*z_0)$
    \STATE $\*z_{T_f} = \text{NF}(\*z_0)$
    \STATE $\*M_h = \*H * (\*M_w \odot \text{reshape}(\*z_{T_f}, [1, 1, D_f]))$
    \STATE $\*V_h = \*H^2 * \!\Sigma_w$
    \STATE $\*E \sim \mathcal{N}(0, 1)$
    \STATE return $\*M_h + \sqrt{\*V_h} \odot \*E$
\end{algorithmic}
\end{algorithm}

For the normalizing flow of $q(\*z)$ we will use the masked RealNVP~\cite{dinh2016density} using the numerically stable updates introduced in Inverse Autoregressive Flow (IAF)~\cite{kingma2016improving}:
\begin{align}
\*m \sim \text{Bern}(0.5); & \qquad \*h = \tanh(f(\*m \odot \*z_t))\nonumber\\
\!\mu = g(\*h); & \qquad \!\sigma = \sigma(k(\*h))\nonumber\\
\*z_{t+1} = \*m \odot \*z_t + & (1 - \*m) \odot (\*z_t \odot \!\sigma + (1 - \!\sigma) \odot \!\mu)\label{eq:nf_bnn}\\
\log \bigg|\frac{\partial\*z_{t+1}}{\partial\*z_t}\bigg| & = (1 - \*m)^T \log\!\sigma,\nonumber
\end{align}
where $\odot$ corresponds to element-wise multiplication, $\sigma(\cdot)$ is the sigmoid function\footnote{$f(x) = \frac{1}{1 + \exp(-x)}$} and $f(\cdot), g(\cdot), k(\cdot)$ are linear mappings. We resampled the mask $\*m$
every time in order to avoid a specific splitting over the dimensions of $\*z$. For the starting point of the flow $q(\*z_0)$ we used a simple fully factorized Gaussian and we will refer to the final iterate as $\*z_{T_f}$.

\subsection{Bounding the entropy}
\label{sec:bound_entr}
Unfortunately, parametrizing the posterior distribution as eq.~\ref{eq:mixture} makes the lower bound intractable as generally we do not have a closed form density function for $q(\*W)$. This makes the calculation of the entropy $ - \E_{q(\*W)}[\log q(\*W)]$ challenging. Fortunately we can make the lower bound tractable again by further lower bounding the entropy in terms of an auxiliary distribution $r(\*z | \*W)$~\citep{agakov2004auxiliary,salimans2013fixed,salimans2015markov,ranganath2015hierarchical,maaloe2016auxiliary}. This can be seen as if we are performing variational inference on the augmented probability space $p(\mathcal{D}, \*W_{1:L}, \*z_{1:L})$, that maintains the same true posterior distribution $p(\*W|\mathcal{D})$ (as we can always marginalize out $r(\*z|\*W)$ to obtain the original model). The lower bound in this case becomes:
\begin{align}
\mathcal{L(\phi, \theta)} & = \E_{q_\phi(\*z_{1:L}, \*W_{1:L})}\big[\log p(\*y | \*x, \*W_{1:L}, \*z_{1:L}) + \nonumber\\& + \log p(\*W_{1:L}) + \log r_\theta(\*z_{1:L}|\*W_{1:L}) - \nonumber \\ & -\log q_\phi(\*W_{1:L} | \*z_{1:L}) - \log q_\phi(\*z_{1:L})\big],\label{eq:bound_paper}
\end{align}
where $\theta$ are the parameters of the auxiliary distribution $r(\cdot)$. This bound is looser than the previous bound, however the extra flexibility of $q(\*W)$ can compensate and allow for a tighter bound. Furthermore, the tightness of the bound also depends on the ability of $r(\*z|\*W)$ to approximate the ``auxiliary" posterior distribution $q(\*z|\*W) = \frac{q(\*W|\*z)q(\*z)}{q(\*W)}$. Therefore, to allow for a flexible $r(\*z|\*W)$ we will follow~\cite{ranganath2015hierarchical} and we will parametrize it with inverse normalizing flows as follows: 
\begin{align}
r(\*z_{T_b}|\*W) & = \prod_{i=1}^{D_z}\mathcal{N}(\tilde{\mu}_i, \tilde{\sigma}^2_i),
\end{align}
where for fully connected layers we have that:
\begin{align}
\tilde{\mu}_i 	& = \big(\*b_1 \otimes \text{tanh}(\*c^T\*W)\big)(\*1\odot D_{out}^{-1})\label{eq:aux_model1}\\
\tilde{\sigma}_i & = \sigma\bigg(\big(\*b_2 \otimes \text{tanh}(\*c^T\*W)\big)(\*1\odot D_{out}^{-1})\bigg),\label{eq:aux_model2}
\end{align}
and for convolutional:
\begin{align}
\tilde{\mu}_i 	& = \big(\text{tanh}(\text{mat}(\*W)\*c)\otimes\*b_1 \big)(\*1\odot (D_{h}D_{w})^{-1})\label{eq:aux_model3}\\
\tilde{\sigma}_i & = \sigma\bigg(\big(\text{tanh}(\text{mat}(\*W)\*c) \otimes \*b_2\big)(\*1\odot (D_{h}D_{w})^{-1})\bigg),\label{eq:aux_model4}
\end{align}
where $\*b_1,\*b_2,\*c$ are trainable vectors that have the same dimensionality as $\*z$, $D_z$, $\*1$ corresponds to a vector of 1s, $\otimes$ corresponds to the outer product and $\text{mat}(\cdot)$ corresponds to the matricization\footnote{Converting the multidimensional tensor to a matrix.} operator. The $\*z_{T_b}$ variable corresponds to the fully factorized variable that is transformed by a normalizing flow to $\*z_{T_f}$ or else the variable obtained by the inverse normalizing flow, $\*z_{T_b} = \text{NF}^{-1}(\*z_{T_f})$. We will parametrize this inverse directly with the procedure described at eq.~\ref{eq:nf_bnn}. Notice that we can employ local reparametrizations also in eq.~\ref{eq:aux_model1},\ref{eq:aux_model2},\ref{eq:aux_model3},\ref{eq:aux_model4}, so as to avoid sampling the, potentially big, matrix $\*W$. With the standard normal prior and the fully factorized Gaussian posterior of eq.~\ref{eq:base_density_ff} the KL-divergence between the prior and the posterior can be computed as follows:
\begin{align}
&- KL(q(\*W)||p(\*W)) =\nonumber\\&\qquad=\E_{q(\*W, \*z_T)}[-KL(q(\*W|\*z_{T_f})||p(\*W)) + \nonumber\\&\qquad +\log r(\*z_{T_f}|\*W) - \log q(\*z_{T_f})],
\end{align}
where each of the terms corresponds to:
\begin{align}
&-KL(q(\*W|\*z_{T_f})||p(\*W)) = \nonumber\\&\qquad = \frac{1}{2}\sum_{i,j}(- \log\sigma^2_{i,j} + \sigma^2_{i,j} + z^2_{T_{f_i}}\mu_{i,j}^2 - 1)\\
&\log r(\*z_{T_f}|\*W) = \log r(\*z_{T_b}|\*W) + \sum_{t=T_f}^{T_f + T_b}\log\bigg|\frac{\partial\*z_{t+1}}{\partial\*z_t}\bigg|\\
& \log q(\*z_{T_f}) = \log q(\*z_0) - \sum_{t=1}^{T_f}\log\bigg|\frac{\partial\*z_{t+1}}{\partial\*z_t}\bigg|.
\end{align} 

It should be noted that this bound is a generalization of the bound proposed by~\cite{gal2015dropout}. We can arrive at the bound of~\cite{gal2015dropout} if we trivially parametrize the auxiliary model $r(\*z|\*W) = q(\*z)$ (which provides a less tight bound~\cite{ranganath2015hierarchical}) use a standard normal prior for $\*W$, a Bernoulli $q(\*z)$ with probability of success $\pi$ and then let the variance of our conditional Gaussian $q(\*W|\*z)$ go to zero. This will result into the lower bound being infinite due to the log of the variances; nevertheless since we are not optimizing over $\sigma$ we can simply disregard those terms. After a little bit of algebra we can show that the only term that will remain in the KL-divergence between $q(\*W)$ and $p(\*W)$ will be the expectation of the trace of the square of the mean matrix\footnote{The matrix that has $\*M[i,j] = \mu_{ij}$}, i.e. $\E_{q(\*z)}[\frac{1}{2}\tr((\diag(\*z)\*M))^T(\diag(\*z)\*M))] = \frac{\pi}{2} \|\*M\|_2^2$, with $1 - \pi$ being the dropout rate.

We also found that in general it is beneficial to ``constrain" the standard deviations $\sigma_{ij}$ of the conditional Gaussian posterior $q(\*W|\*z)$ during the forward pass for the computation of the likelihood to a lower than the true range, e.g. $[0, \alpha]$ instead of the $[0, 1]$ we have with a standard normal prior. This results into a small bias and a looser lower bound, however it helps in avoiding bad local minima in the variational objective. This is akin to the free bits objective described at~\cite{kingma2016improving}.

\section{Related work}
\label{related}
Approximate inference for Bayesian neural networks has been pioneered by~\cite{mackay1992practical} and~\cite{neal1995bayesian}. Laplace approximation~\citep{mackay1992practical} provides a deterministic approximation to the posterior that is easy to obtain; it is a Gaussian centered at the MAP estimate of the parameters with a covariance determined by the inverse of the Hessian of the log-likelihood. Despite the fact that it is straightforward to implement, its scalability is limited unless approximations are made, which generally reduces performance. Hamiltonian Monte Carlo~\citep{neal1995bayesian} is so far the golden standard for approximate Bayesian inference; nevertheless it is also not scalable to large networks and datasets due to the fact that we have to explicitly store the  samples from the posterior. Furthermore as it is an MCMC method, assessing convergence is non trivial. Nevertheless there is interesting work that tries to improve upon those issues with stochastic gradient MCMC~\citep{chen2014stochastic} and distillation methods~\citep{korattikara2015bayesian}.

Deterministic methods for approximate inference in Bayesian neural networks have recently attained much attention. One of the first applications of variational inference in neural networks was in~\cite{peterson1987mean} and~\cite{hinton1993keeping}. More recently~\cite{graves2011practical} proposed a practical method for variational inference in this setting with a simple (but biased) estimator for a fully factorized posterior distribution.~\cite{blundell2015weight} improved upon this work with the unbiased estimator from~\cite{kingma2013auto} and a scale mixture prior.~\cite{hernandez2015probabilistic} proposed to use Expectation Propagation~\citep{minka2001expectation} with fully factorized posteriors and showed good results on regression tasks.~\cite{kingma2015variational} showed how Gaussian dropout can be interpreted as performing approximate inference with log-uniform priors, multiplicative Gaussian posteriors and local reparametrizations, thus allowing straightforward learning of the dropout rates. Similarly~\cite{gal2015dropout} showed interesting connections between Bernoulli Dropout~\citep{srivastava2014dropout} networks and approximate Bayesian inference in deep Gaussian Processes~\citep{damianou2012deep} thus allowing the extraction of uncertainties in a principled way. Similarly~\cite{louizos2016structured} arrived at the same result through structured posterior approximations via matrix Gaussians and local reparametrizations~\citep{kingma2015variational}. 

It should also be mentioned that uncertainty estimation in neural networks can also be performed without the Bayesian paradigm; frequentist methods such as Bootstrap~\cite{osband2016deep} and ensembles~\cite{lakshminarayanan2016simple} have shown that in certain scenarios they can provide reasonable confidence intervals. 

\section{Experiments}
\label{experiments}
All of the experiments were coded in Tensorflow~\citep{abadi2016tensorflow} and optimization was done with Adam~\citep{kingma2014adam} using the default hyper-parameters. We used the LeNet 5\footnote{The version from Caffe.}~\cite{lecun1998gradient} convolutional architecture with ReLU~\cite{nair2010rectified} nonlinearities. The means $\*M$ of the conditional Gaussian $q(\*W|\*z)$ were initialized with the scheme proposed in~\cite{he2015delving}, whereas the log of the variances were initialized by sampling from $\mathcal{N}(-9, 0.001)$. Unless explicitly mentioned otherwise we use flows of length two for $q(\*z)$ and $r(\*z|\*W)$ with 50 hidden units for each step of the flow of $q(\*z)$ and 100 hidden units for each step of the flow of $r(\*z|\*W)$. We used 100 posterior samples to estimate the predictive distribution for all of the models during testing and 1 posterior sample during training.

\begin{table}[htb]
\centering 
\caption{Models considered in this paper. Dropout corresponds to the model used in~\cite{gal2015bayesian}, Deep Ensemble to the model used in~\cite{lakshminarayanan2016simple}, FFG to the Bayesian neural network employed in~\cite{blundell2015weight}, FFLU to the Bayesian neural network used in~\cite{kingma2015variational,2017arXiv170105369M} with the additive parametrization of~\cite{2017arXiv170105369M} and MNFG corresponds to the proposed variational approximation. It should be noted that Deep Ensembles use adversarial training~\cite{goodfellow2014explaining}.}
\label{tab:models_trained}
\resizebox{ \columnwidth}{!}{%
\begin{tabular}{l|c|c}
\textbf{Name} & \textbf{Prior} & \textbf{Posterior} \\\hline
\textbf{L2} & $\mathcal{N}(\*0, \*I)$ & delta peak \\
\textbf{Dropout} & $\mathcal{N}(\*0, \*I) $ & mixture of zero and delta peaks \\
\textbf{D. Ensem.} & - & mixture of peaks \\
\textbf{FFG} & $\mathcal{N}(\*0, \*I)$& fully factorized additive Gaussian \\
\textbf{FFLU} & $\log (|\*W|)$ = c& fully factorized additive Gaussian \\
\textbf{MNFG} & $\mathcal{N}(\*0, \*I)$ & multiplicative normalizing flows \\
\end{tabular} %
}
\end{table}

\subsection{Predictive performance and uncertainty}
\paragraph{MNIST} We trained on MNIST LeNet architectures using the priors and posteriors described at Table~\ref{tab:models_trained}. We trained Dropout with the way described at~\cite{gal2015bayesian} using 0.5 for the dropout rate and for Deep Ensembles~\cite{lakshminarayanan2016simple} we used 10 members and $\epsilon=.25$ for the adversarial example generation. For the models with the Gaussian prior we constrained the standard deviation of the conditional posterior to be $\leq .5$ during the forward pass. The classification performance of each model can be seen at Table~\ref{tab:class_results}; while our overall focus is not classification accuracy per se, we see that with the MNF posteriors we improve upon mean field reaching similar accuracies with Deep Ensembles.

\begin{figure}[htb]
\centering
\includegraphics[width=.88\columnwidth]{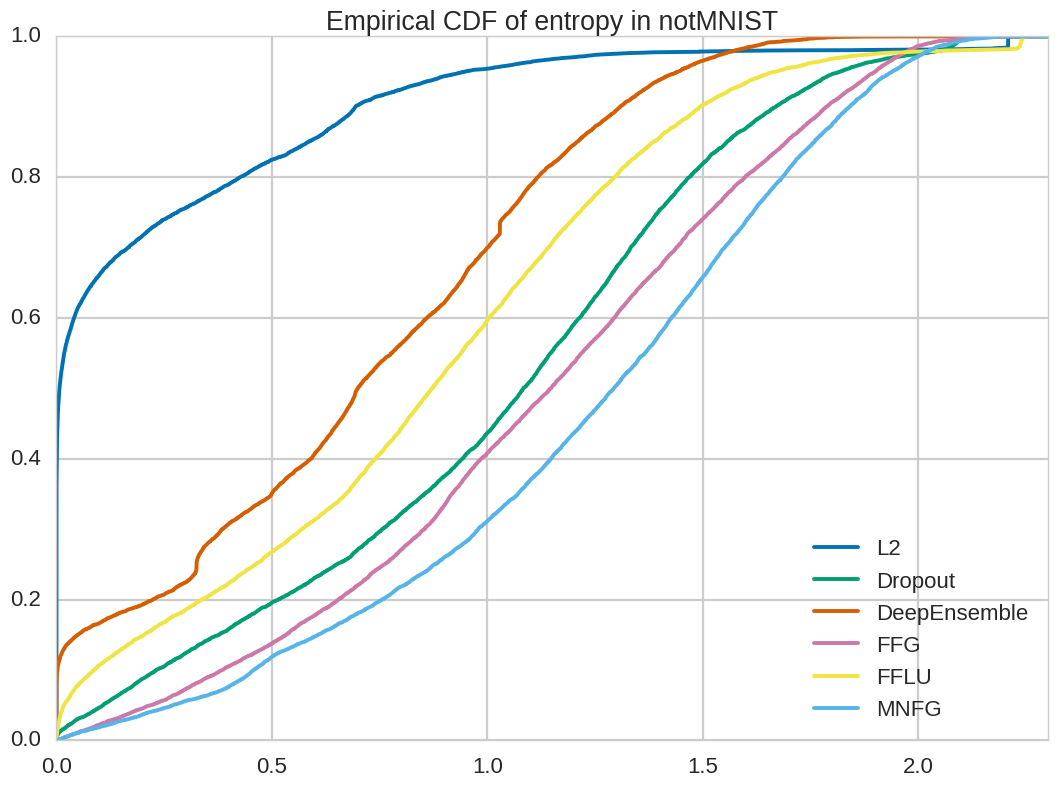}
\caption{Empirical CDF for the entropy of the predictive distributions on notMNIST.}
\label{fig:entr_notmnist}
\end{figure}

\paragraph{notMNIST} To evaluate the predictive uncertainties of each model we performed the task described at~\cite{lakshminarayanan2016simple}; we estimated the entropy of the predictive distributions on notMNIST\footnote{Can be found at \url{http://yaroslavvb.blogspot.co.uk/2011/09/notmnist-dataset.html}} from the LeNet architectures trained on MNIST. Since we a-priori know that none of the notMNIST classes correspond to a trained class (since they are letters and not digits) the ideal predictive distribution is uniform over the MNIST digits, i.e. a maximum entropy distribution. Contrary to~\cite{lakshminarayanan2016simple} we do not plot the histogram of the entropies across the images but we instead use the empirical CDF, which we think is more informative. Curves that are closer to the bottom right part of the plot are preferable, as it denotes that the probability of observing a high confidence prediction is low. At Figure~\ref{fig:entr_notmnist} we show the empirical CDF over the range of possible entropies, $[0, 2.5]$, for all of the models. 

It is clear from the plot that the uncertainty estimates from MNFs are better than the other approaches, since the probability of a low entropy prediction is overall lower. The network trained with just weight decay was, as expected, the most overconfident with an almost zero median entropy while Dropout seems to be in the middle ground. The Bayesian neural net with the log-uniform prior also showed overconfidence in this task; we hypothesize that this is due to the induced sparsity~\citep{2017arXiv170105369M} which results into the pruning of almost all irrelevant sources of variation in the parameters thus not providing enough variability to allow for uncertainty in the predictions.  The sparsity levels\footnote{Computed by pruning weights where $\log\sigma^2 - \log\mu^2 \geq 5$~\cite{2017arXiv170105369M}.} are 62\%, 95.2\% for the two convolutional layers and 99.5\%, 93.3\% for the two fully connected. Similar effects would probably be also observed if we optimized the dropout rates for Dropout. The only source of randomness in the neural network is from the Bernoulli random variables (r.v.) $z$. By employing the Central Limit Theorem\footnote{Assuming that the network is wide enough.} we can express the distribution of the activations as a Gaussian~\cite{wang2013fast} with variance affected by the variance of the Bernoulli r.v., $\mathbb{V}(z) = \pi(1 - \pi)$. The maximum variance of the Bernoulli r.v. is when $\pi = 0.5$, therefore any tuning of the Dropout rate will result into a decrease in the variance of the r.v. and therefore a decrease in the variance of the Gaussian at the hidden units. This will subsequently lead into less predictive variance and more confidence. 

Finally, whereas it was shown at~\cite{lakshminarayanan2016simple} that Deep Ensembles provide good uncertainty estimates (better than Dropout) on this task using fully connected networks, this result did not seem to apply for the LeNet architecture we considered. We hypothesize that they are sensitive to the hyperparameters (e.g. adversarial noise, number of members in the ensemble) and it requires more tuning in order to improve upon Dropout on this architecture.

\begin{figure}[htb]
\centering
\includegraphics[width=.88\columnwidth]{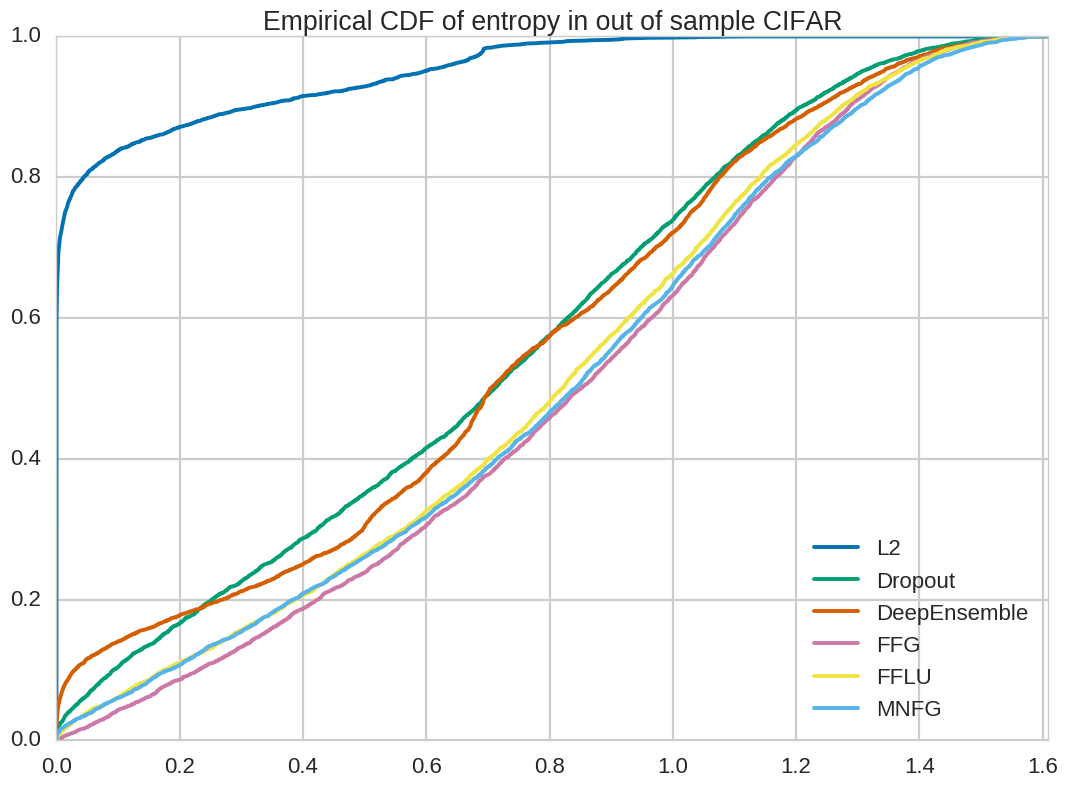}
\caption{Empirical CDF for the entropy of the predictive distributions on the 5 hidden classes from CIFAR 10.}
\label{fig:entr_cifar}
\end{figure}

\paragraph{CIFAR 10} We performed a similar experiment on CIFAR 10. To artificially create the "unobserved class" scenario, we hid 5 of the labels (dog, frog, horse, ship, truck) and trained on the rest (airplane, automobile, bird, cat, deer). For this task we used the larger LeNet architecture\footnote{192 filters at each convolutional layer and 1000 hidden units for the fully connected layer.} described at~\cite{gal2015bayesian}. For the models with the Gaussian prior we similarly constrained the standard deviation during the forward pass to be $\leq .4$. For Deep Ensembles we used five members with $\epsilon=.1$ for the adversarial example generation. The predictive performance on these five classes can be seen in Table~\ref{tab:class_results}, with Dropout and MNFs achieving the overall better accuracies. We subsequently measured the entropy of the predictive distribution on the classes that were hidden, with the resulting empirical CDFs visualized in Figure~\ref{fig:entr_cifar}.

We similarly observe that the network with just weight decay was the most overconfident. Furthermore, Deep Ensembles and Dropout had similar uncertainties, with Deep Ensembles having lower accuracy on the observed classes. The networks with the Gaussian priors also had similar uncertainty with the network with the log uniform prior, nevertheless the MNF posterior had much better accuracy on the observed classes. The sparsity levels for the network with the log-uniform prior now were 94.9\%, 99.8\% for the convolutional layers and 99.9\%, 92.7\% for the fully connected. Overall, the network with the MNF posteriors seem to provide the better trade-off in uncertainty and accuracy on the observed classes.

\begin{table}[htb]
\centering 
\caption{Test errors (\%) with the LeNet architecture on MNIST and the first five classes of CIFAR 10.}
\label{tab:class_results}
\resizebox{\columnwidth}{!}{%
\begin{tabular}{l|c|c|c|c|c|c}
\textbf{Dataset} & \textbf{L2} & \textbf{Dropout} & \textbf{D.Ensem.} & \textbf{FFG} & \textbf{FFLU} & \textbf{MNFG}\\\hline
\textbf{MNIST} & 0.6 & 0.5 & 0.7 & 0.9 & 0.9 & 0.7\\
\textbf{CIFAR 5} & 24 & 16 & 21 & 22 & 23 & 16\\
\end{tabular}}
\end{table}

\subsection{Accuracy and uncertainty on adversarial examples}
We also measure how robust our models and uncertainties are against adversarial examples~\cite{szegedy2013intriguing,goodfellow2014explaining} by generating examples using the fast sign method~\cite{goodfellow2014explaining} for each of the previously trained architectures using Cleverhans~\cite{papernot2016cleverhans}. For this task we do not include Deep Ensembles as they are trained on adversarial examples.  

\begin{figure}[htb]
\centering
\includegraphics[width=.88\columnwidth]{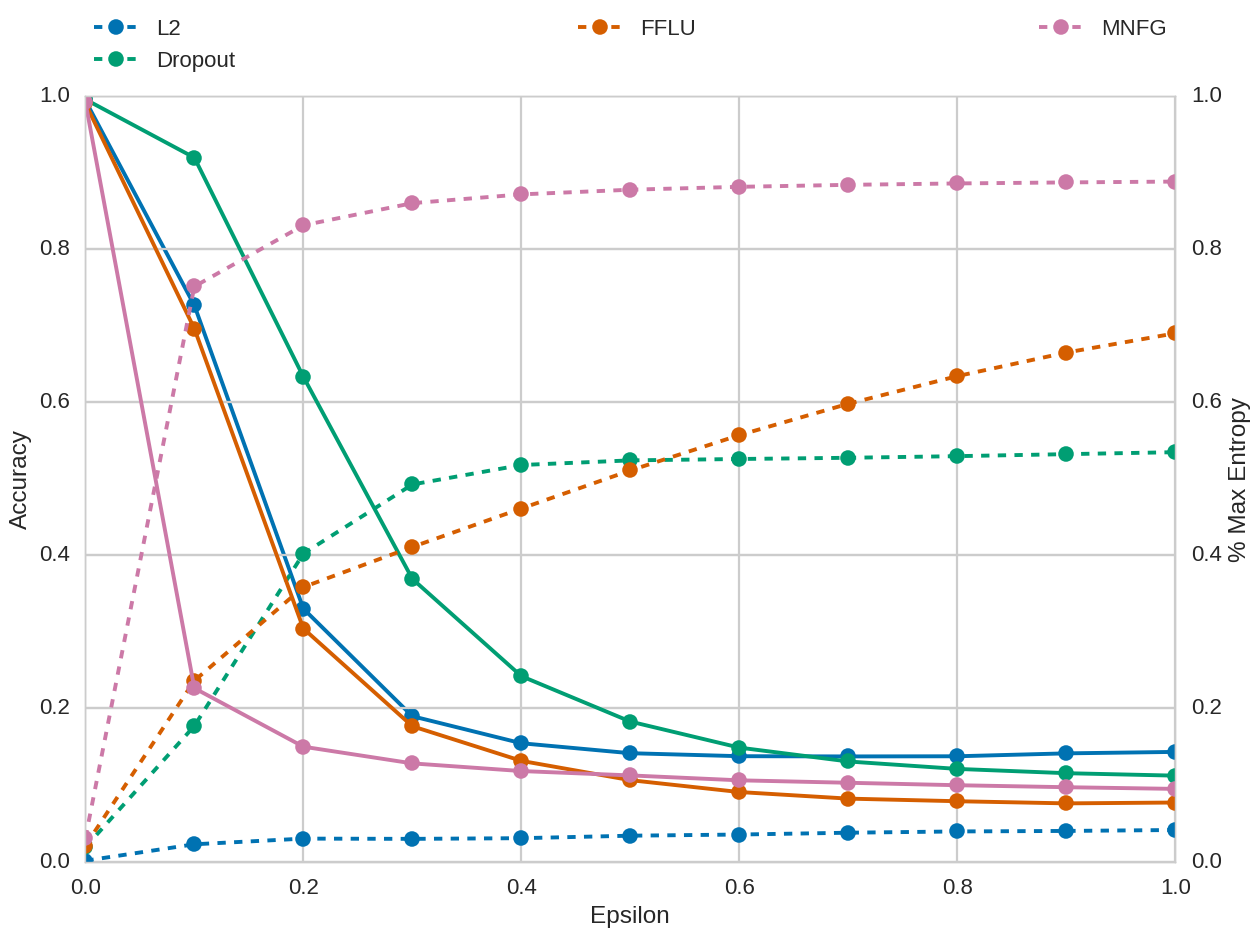}
\caption{Accuracy (solid) vs entropy (dashed) as a function of the adversarial perturbation $\epsilon$ on MNIST.}
\label{fig:advers_mnist}
\end{figure}

\paragraph{MNIST} On this scenario we observe interesting results if we plot the change in accuracy and entropy by varying the magnitude of the adversarial perturbation. The resulting plot can be seen in Figure~\ref{fig:advers_mnist}. Overall Dropout seems to have better accuracies on adversarial examples; nevertheless, those come at an "overconfident" price since the entropy of the predictive distributions is quite low thus resulting into predictions that have, on average, above 0.7 probability for the dominant class. This is in contrast with MNFs; while the accuracy almost immediately drops close to random, the uncertainty simultaneously increases to almost maximum entropy. This implies that the predictive distribution is more or less uniform over those examples. So despite the fact that our model cannot overcome adversarial examples at least it ``knows that it doesn't know".

\begin{figure}[htb]
\centering
\includegraphics[width=.88\columnwidth]{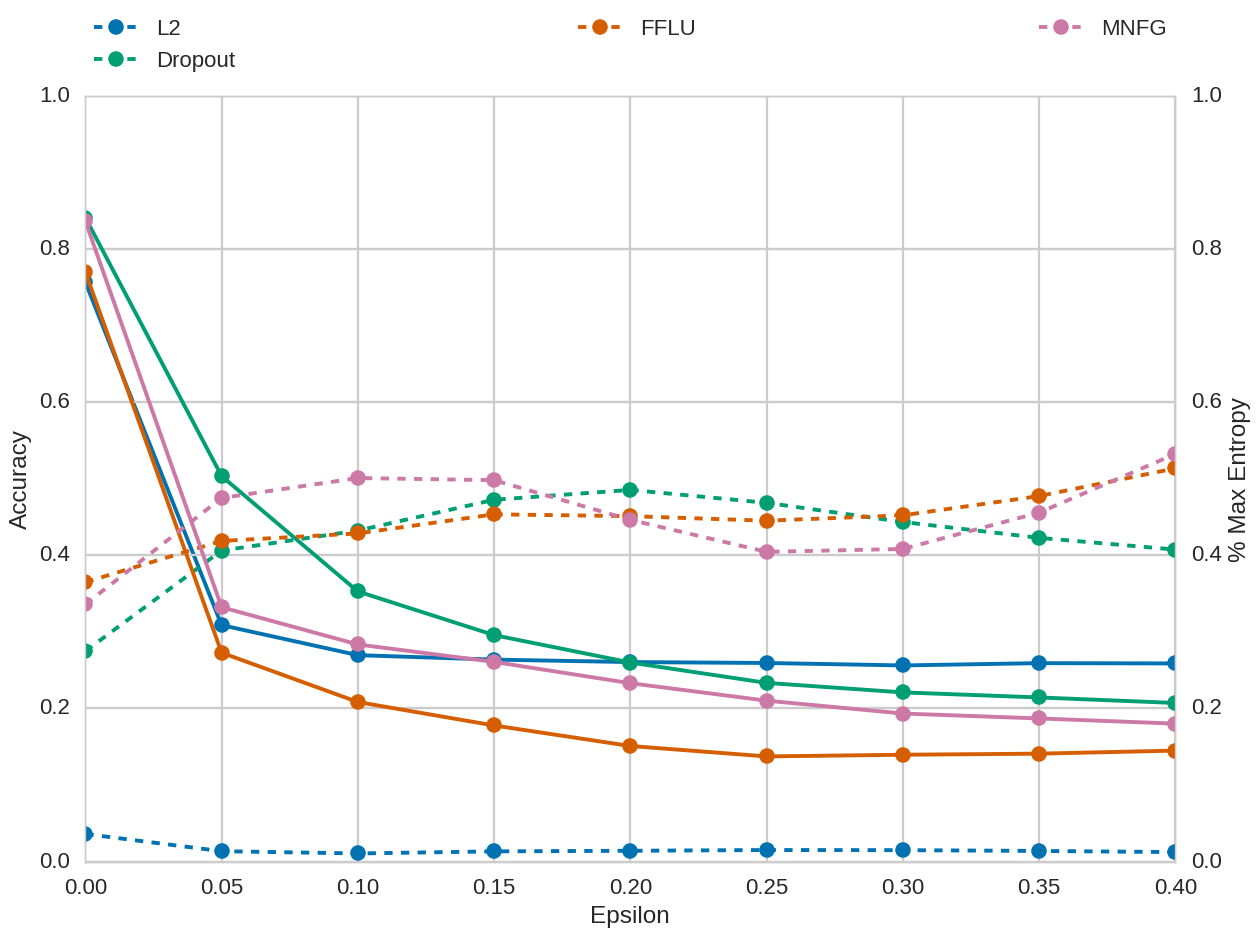}
\caption{Accuracy (solid) vs entropy (dashed) as a function of the adversarial perturbation $\epsilon$ on CIFAR 10 (on the first 5 classes).}
\label{fig:advers_cifar}
\end{figure}

\paragraph{CIFAR} We performed the same experiment also on the five class subset of CIFAR 10. The results can be seen in Figure~\ref{fig:advers_cifar}. Here we however observe a different picture, compared to MNIST, since all of the methods experienced overconfidence.  We hypothesize that adversarial examples are harder to escape and be uncertain about in this dataset, due to the higher dimensionality, and therefore further investigation is needed.

\begin{figure*}[htb]
\centering
\begin{subfigure}{.25\textwidth}
  \centering
  \includegraphics[width=\linewidth]{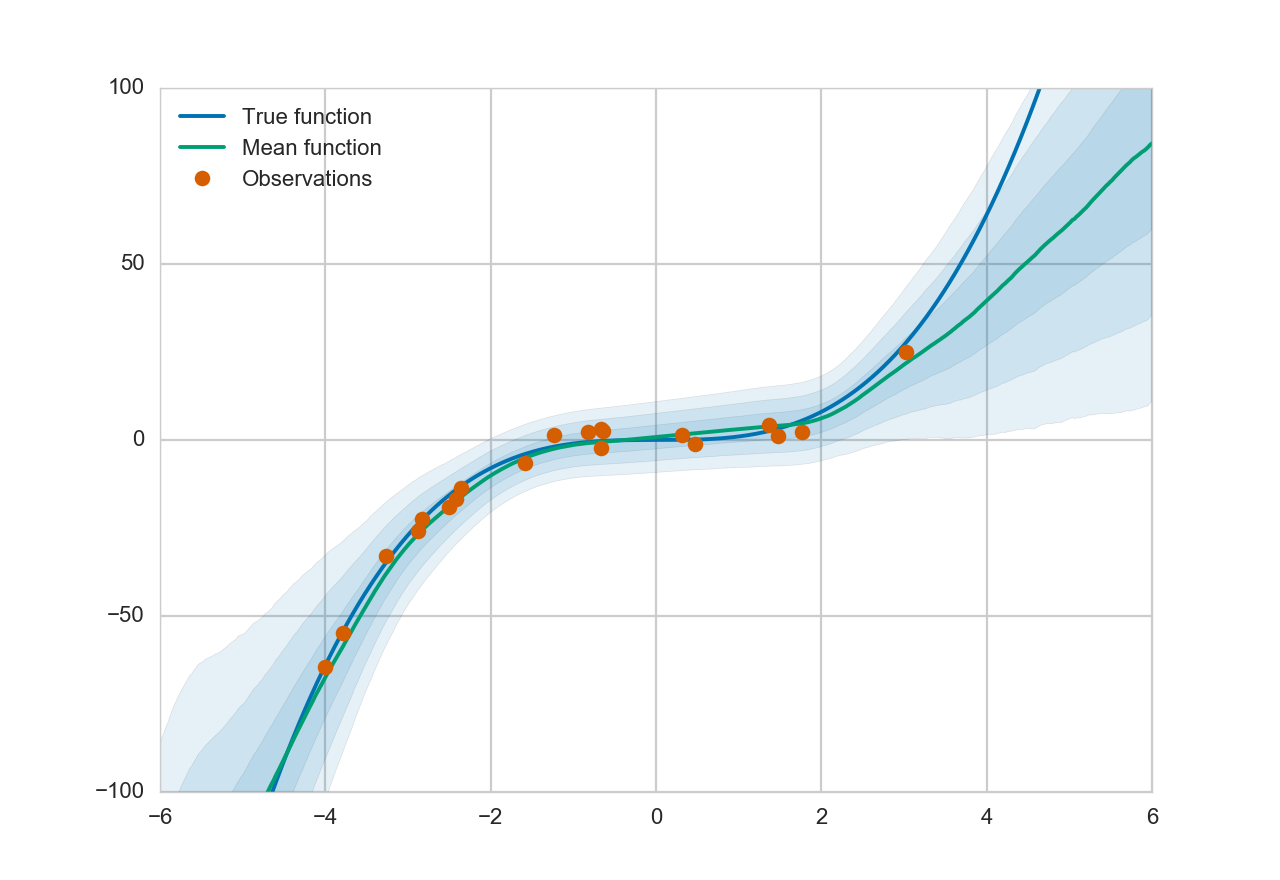}
  \caption{Dropout $\pi = 0.5$}
  \label{fig:dropout_toy}
\end{subfigure}%
\begin{subfigure}{.25\textwidth}
  \centering
  \includegraphics[width=\linewidth]{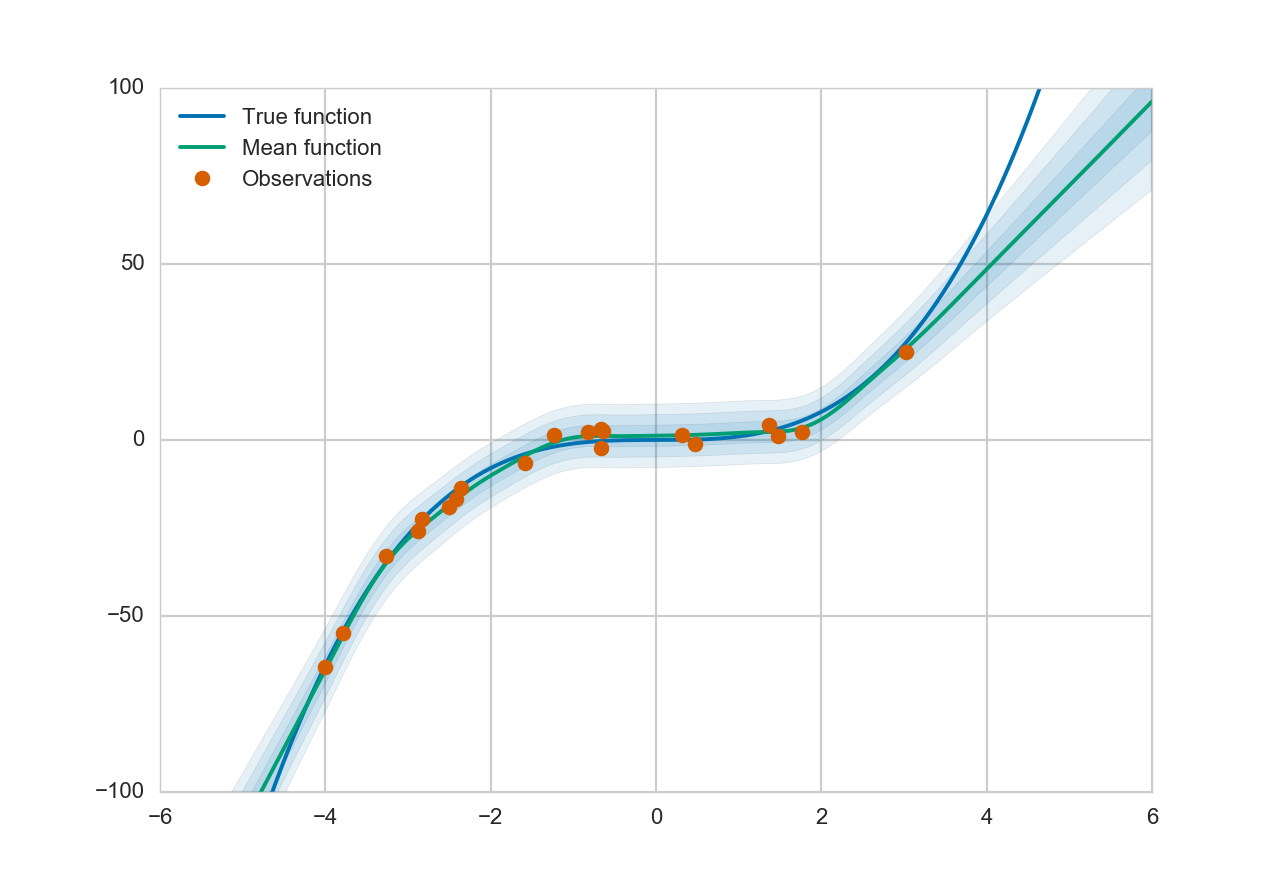}
  \caption{Dropout learned $\pi$}
  \label{fig:vardrop_toy}
\end{subfigure}%
\begin{subfigure}{.25\textwidth}
  \centering
  \includegraphics[width=\linewidth]{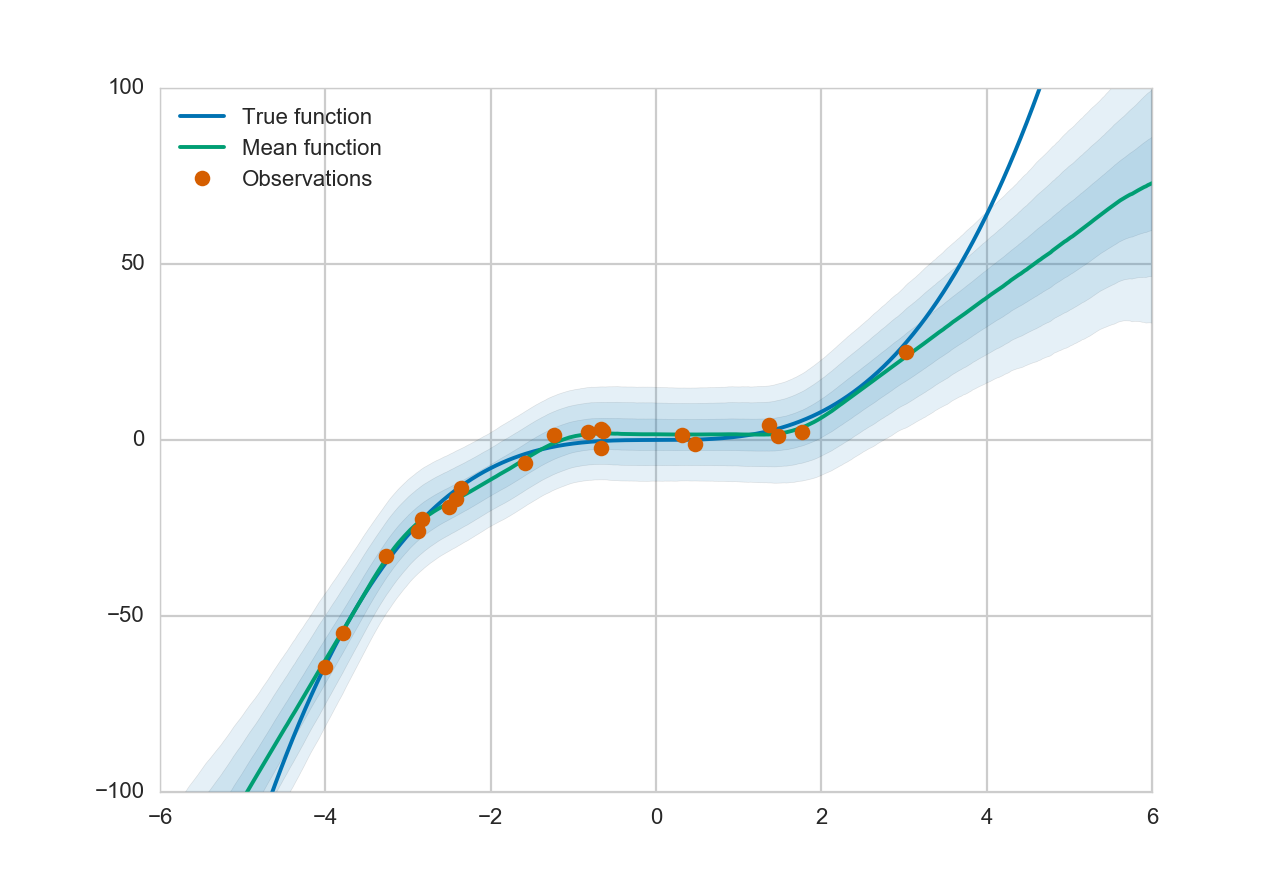}
  \caption{FFLU}
  \label{fig:vardrop_add_toy}
\end{subfigure}%
\begin{subfigure}{.25\textwidth}
  \centering
  \includegraphics[width=\linewidth]{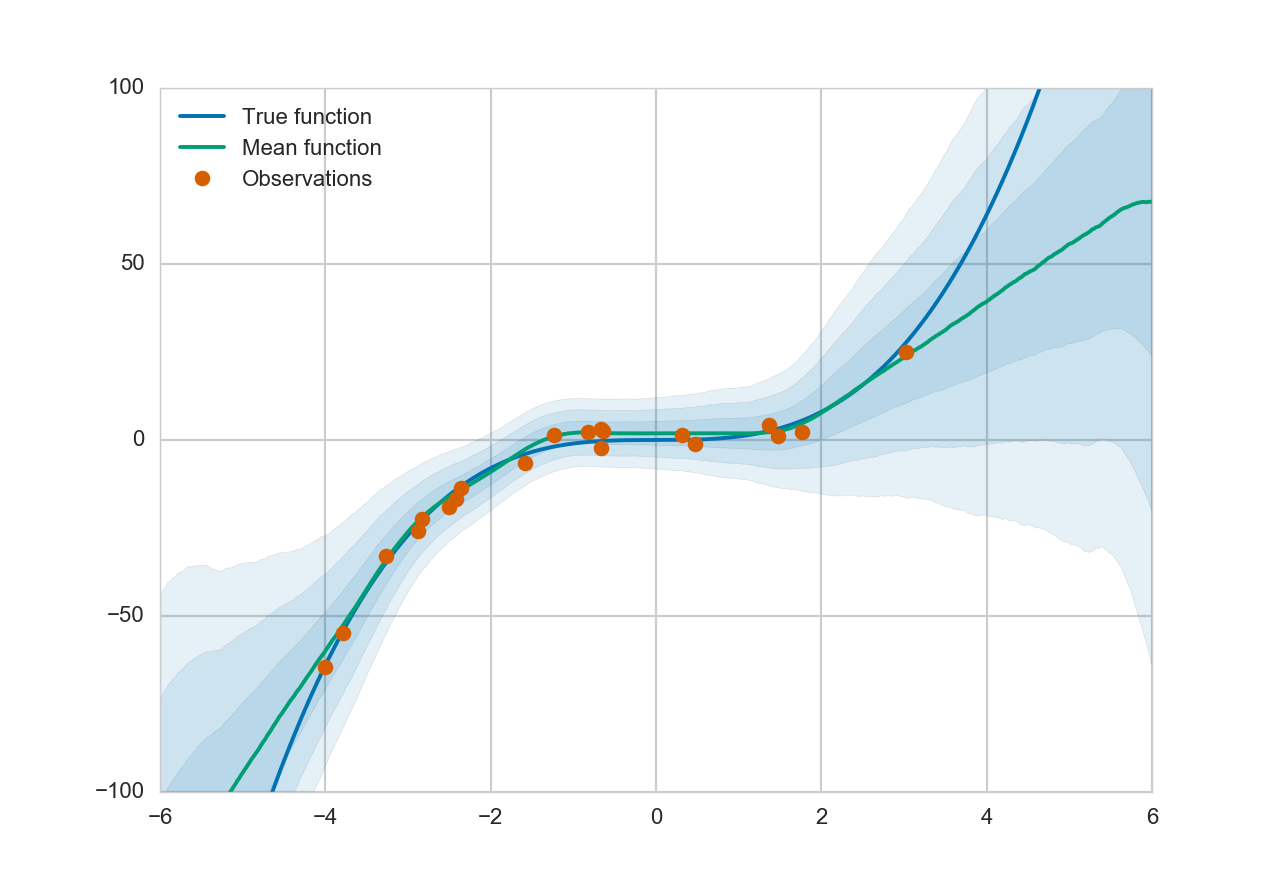}
  \caption{MNFG}
  \label{fig:mnf_toy}
\end{subfigure}
\caption{Predictive distributions for the toy dataset. Blue areas correspond to $\pm$3 standard deviations around the mean.}
\label{fig:toy_reg_fig}
\end{figure*}

\subsection{Regression on toy dataset}
For the final experiment we visualize the predictive distributions obtained with the different models on the toy regression task introduced at~\cite{hernandez2015probabilistic}. We generated 20 training inputs from $\mathcal{U}[-4, 4]$ and then obtained the corresponding targets via $y = x^3 + \epsilon$, where $\epsilon \sim \mathcal{N}(0, 9)$. We fixed the likelihood noise to its true value and then fitted a Dropout network with $\pi = 0.5$ for the hidden layer\footnote{No Dropout was used for the input layer since it is 1-dimensional.}, an FFLU network and an MNFG. We also fitted a Dropout network where we also learned the dropout probability $\pi$ of the hidden layer according to the bound described at section~\ref{sec:bound_entr} (which is equivalent to the one described at~\cite{gal2015dropout}) using REINFORCE~\cite{williams1992simple} and a global baseline~\cite{mnih2014neural}. The resulting predictive distributions can be seen at Figure~\ref{fig:toy_reg_fig}.

As we can observe, MNF posteriors provide more realistic predictive distributions, closer to the true posterior (which can be seen at~\cite{hernandez2015probabilistic}) and with the network being more uncertain on areas where we do not observed any data. The uncertainties obtained by Dropout with fixed $\pi=0.5$ did not diverge as much in those areas but overall they were better compared to the uncertainties obtained with FFLU. We could probably attribute the latter to the sparsification of the network since 95\% and 44\% of the parameters were pruned for each layer respectively. 

Interestingly the uncertainties obtained with the network with the learned Dropout probability were the most ``overfitted". This might suggest that Dropout uncertainty is probably not a good posterior approximation since by optimizing the dropout rates we do not seem to move closer to the true posterior predictive distribution. This is in contrast with MNFs; they are flexible enough to allow for optimizing all of their parameters in a way that does better approximate the true posterior distribution. This result also empirically verifies the claim we previously made; by learning the dropout rates the entropy of the posterior predictive will decrease thus resulting into more overconfident predictions.

\section{Conclusion}
\label{conclusion}
We introduce multiplicative normalizing flows (MNFs); a family of approximate posteriors for the parameters of a variational Bayesian neural network. We have shown that through this approximation we can significantly improve upon mean field on both predictive performance as well as predictive uncertainty. We compared our uncertainty on notMNIST and CIFAR with Dropout~\cite{srivastava2014dropout,gal2015dropout} and Deep Ensembles~\cite{lakshminarayanan2016simple} using convolutional architectures and found that MNFs achieve more realistic uncertainties while providing predictive capabilities on par with Dropout. We suspect that the predictive capabilities of MNFs can be further improved through more appropriate optimizers that avoid the bad local minima in the variational objective. Finally, we also highlighted limitations of Dropout approximations and empirically showed that MNFs can overcome them. 

There are a couple of promising directions for future research. One avenue would be to explore how much can MNFs sparsify and compress neural networks under either sparsity inducing priors, such as the log-uniform prior~\cite{kingma2015variational,2017arXiv170105369M}, or empirical priors~\cite{ullrich2017soft}. Another promising direction is that of designing better priors for Bayesian neural networks. For example~\cite{neal1995bayesian} has identified limitations of Gaussian priors and proposes alternative priors such as the Cauchy. Furthermore, the prior over the parameters also affects the type of uncertainty we get in our predictions; for instance we observed in our experiments a significant difference in uncertainty between Gaussian and log-uniform priors. Since different problems require different types of uncertainty it makes sense to choose the prior accordingly, e.g. use an informative prior so as to alleviate adversarial examples.

\section*{Acknowledgements} 
We would like to thank Klamer Schutte, Matthias Reisser and Karen Ullrich for valuable feedback. This research is supported by TNO, NWO and Google.

\bibliography{example_paper}
\bibliographystyle{icmlnew}

\newpage
\clearpage
\appendix
\section{Memorization capabilities}
As it was shown in~\cite{zhang2016understanding}, deep neural networks can exhibit memorization, even with random labels. Therefore deep neural networks could perfectly fit the training data while having random chance accuracy on the test data, even with Dropout or weight decay regularization.~\cite{2017arXiv170105369M} instead showed that by employing Sparse Variational Dropout this phenomenon did not appear, thus resulting into the network pruning everything and having random chance accuracy on both training and test sets. We similarly show here that with Gaussian priors and MNF posteriors we also have random chance accuracy on both train and test sets. This suggests that it is proper Bayesian inference that penalizes memorization.

\begin{table}[htb]
\centering 
\caption{Accuracy (\%) with the LeNet architecture on MNIST and the first five classes of CIFAR 10 using random labels. Random chance is $11\%$ on MNIST and $20\%$ on CIFAR 5.}
\label{tab:class_memo}
\resizebox{\columnwidth}{!}{%
\begin{tabular}{l|c|c|c|c}
\textbf{Dataset} & \textbf{Dropout} train & \textbf{Dropout} test & \textbf{MNFG} train & \textbf{MNFG} test\\\hline
\textbf{MNIST} & 30 & 11 & 11 & 11 \\
\textbf{CIFAR 5} & 89 & 20 & 20 & 20 \\
\end{tabular}}
\end{table}

\end{document}